\documentclass[11pt,a4paper]{article}
\usepackage{acl2016}
\usepackage{times}
\usepackage{url}
\usepackage{latexsym}
\usepackage{xspace}
\usepackage{booktabs}
\usepackage{color}
\usepackage{graphicx}
\usepackage{tabularx}
\usepackage{tabulary}
\usepackage{enumitem}

\aclfinalcopy %
\def\aclpaperid{637} %

\usepackage{amsfonts}

  {\begin{itemize}[topsep=0pt, partopsep=0pt] %
    \setlength{\itemsep}{0pt}%
    \setlength{\parskip}{0pt}%
    }%
  {\end{itemize}}
  
  \usepackage{color}
  {\begin{enumerate}≈ \footnotesize%
    \setlength{\itemsep}{0pt}%
    \setlength{\parskip}{0pt}%
    }%
  {\end{enumerate}}

\usepackage{microtype}
\usepackage{multirow}
\usepackage{verbatim}
\usepackage{amsmath,amsthm,amssymb}
\usepackage{array}
\usepackage[scaled=0.86]{helvet}
\usepackage{ifthen}
\usepackage{courier}
\usepackage[linesnumbered,vlined,ruled]{algorithm2e}

\newcommand{\captionfonts}{\small}
\makeatletter  %
\long\def\@makecaption#1#2{%
  \vskip\abovecaptionskip
  \sbox\@tempboxa{{\captionfonts #1: #2}}%
  \ifdim \wd\@tempboxa >\hsize
    {\captionfonts #1: #2\par}
  \else
    \hbox to\hsize{\hfil\box\@tempboxa\hfil}%
  \fi
  \vskip\belowcaptionskip}
\makeatother   %

\belowdisplayskip \abovedisplayskip

\newcommand{\reduceVerticalSpace}{
\setlength{\topsep}{0pt}
\setlength{\partopsep}{0pt}
\setlength{\parskip}{0pt}
\setlength{\parsep}{0pt}
\setlength{\itemsep}{0pt}
\setlength{\itemindent}{0pt}
\setlength{\listparindent}{0pt}
}

\newcommand{\bleu}{{{\sc Bleu}}\xspace}

\newcommand{\eos}{{\it EOS}\xspace}
\newcommand{\sts}{{{\textsc{Seq2Seq}}}\xspace}

\newcommand{\Message}{{{\it message}}\xspace}
\newcommand{\Response}{{{\it response}}\xspace}

\newcommand{\User}[1]{{{\it user{#1}}}\xspace}

\title{A Persona-Based Neural Conversation Model}

\author{
Jiwei Li$^{\textnormal {{1}*}}$\hspace{.6cm} %
Michel Galley$^{\textnormal {2}}$\hspace{.6cm}%
Chris Brockett$^{\textnormal {2}}$\\%\!{\textnormal ,}\\
{\bf Georgios P. Spithourakis}$^{\textnormal {3*}}$\hspace{.4cm}%
{\bf Jianfeng Gao}$^{\textnormal {2}}$\hspace{.4cm}%
{\bf Bill Dolan}$^{\textnormal {2}}$
\\[0.4cm]
{$^1$Stanford University, Dept. of Computer Science}\\
{\tt jiweil@stanford.edu} \\
{$^2$Microsoft Research, Redmond}\\
{\tt \{mgalley,chrisbkt,jfgao,billdol\}@microsoft.com}\\
{$^3$University College London, Dept. of Computer Science}\\
{\tt g.spithourakis@cs.ucl.ac.uk}
}

\begin{document}
\maketitle

{\let\thefootnote\relax\footnotetext{* The entirety of this work was conducted at Microsoft.}}

\begin{abstract}
We present persona-based models for handling the issue of speaker consistency in neural response generation. 
A speaker model encodes personas in distributed embeddings that capture individual characteristics such as background information and speaking style.
A dyadic speaker-addressee model captures properties of interactions between two interlocutors.
Our models yield qualitative performance improvements in both perplexity and \bleu scores over baseline sequence-to-sequence models, with similar gains in speaker consistency as measured by human judges. 
\end{abstract}
\section{Introduction}

\label{sec:intro}

As conversational agents gain traction as user interfaces, there has been growing research interest in training naturalistic conversation systems from large volumes of human-to-human interactions \cite{ritter2011data,Sordoni2015,vinyals2015neural,li2015diversity}. 
One major issue for these data-driven systems is their propensity to select the response with greatest likelihood---in effect a consensus response of the humans represented in the training data. Outputs are frequently vague or non-committal \cite{li2015diversity}, and when not, they can be wildly inconsistent, as illustrated in Table \ref{sample:baseline}. 

\begin{table}
\setlength{\tabcolsep}{4pt}
\center
{\small
\begin{tabularx}{\linewidth}{rl}\hline
\Message  & Where do you live now?\\
\Response & I live in Los Angeles.\\
\Message  & In which city do you live now?\\
\Response & I live in Madrid.\\
\Message  & In which country do you live now?\\
\Response & England, you?\\\hline%
\Message  & Where were you born?\\
\Response & I was born in Canada.\\
\Message  & Where are you from?\\
\Response & England, you?\\
\Message  & Where did you grow up?\\
\Response & I grew up in Texas.\\\hline%
\Message  & How old are you?\\
\Response & 16 and you?\\
\Message  & What's your age?\\
\Response & 18.\\\hline%
\Message  & What is your major?\\
\Response & I'm majoring in psychology\\
\Message  & What did you study in college?\\
\Response & English lit. \\\hline%
\end{tabularx}
}
\caption{Inconsistent responses generated by a 4-layer \sts model trained on 25 million Twitter conversation snippets.}
\label{sample:baseline}
\end{table}

In this paper, we address the challenge of consistency and how to endow data-driven systems with the coherent ``persona'' needed to model human-like behavior, whether as personal assistants, personalized avatar-like agents, or game characters.\footnote{\cite{vinyals2015neural} suggest that the lack of a coherent personality makes it impossible for current systems to pass the Turing test.} 
For present purposes, we will define \textsc{persona} as the character that an artificial agent, as actor, plays or performs during conversational interactions.
A~persona can be viewed as a composite of elements of identity (background facts or user profile), language behavior, and interaction style. 
A persona is also adaptive, since an agent may need to present different facets to different human interlocutors depending on the interaction.

Fortunately, neural models
of conversation generation \cite{Sordoni2015,shang2015neural,vinyals2015neural,li2015diversity} provide a straightforward mechanism for incorporating personas as embeddings.
%
%
%
%
\begin{comment}
From JFGao
[085 and 089] We mention that both persona models incorporate persona vectors at “decoding time”. Note that the term “decoding” is ambiguous. It has two senses:
1.	The Seq2Seq model is also called encoder-decoder model, as shown in Figure 1, where the encoding-decoding process could happen either in the process of model training or in the process of generating responses using the trained model.  
2.	As in 4.4. (line 377), “decoding” means applying the trained model to generate responses. 

I think in line 085 and line 089, “decoding” refers to sense 1, not sense 2 (which means “testing time”). So, we need to clarify: 
1.	Seq2Seq is based on an encoder-decoder framework.
2.	Persona vectors are incorporated in the decoder part of the Sea2seq model, as in Eq (4) and Eq (8).
3.	Persona vectors are trained on human-human conversation data.
4.	The trained persona vectors are used at the “testing time” to generate personalized responses.
\end{comment}
We therefore explore two persona models, a single-speaker \textsc{Speaker Model} and a dyadic \textsc{Speaker-Addressee Model}, within a sequence-to-sequence (\sts) framework \cite{sutskever2014sequence}. 
The Speaker Model integrates a speaker-level vector representation into the target part of the \sts model.
Analogously, the Speaker-Addressee model encodes the interaction patterns of two interlocutors by constructing an interaction representation from their individual embeddings and incorporating it into the \sts model. 
These persona vectors are trained on human-human conversation data and used at test time to generate personalized responses.
Our experiments on an open-domain corpus of Twitter conversations and dialog datasets comprising TV series scripts show that leveraging persona vectors can improve relative performance up to $20\%$ in \bleu score and $12\%$ in perplexity, with a commensurate gain in consistency as judged by human annotators.

\section{Related Work}
\label{sec:related}
This work follows the line of investigation initiated by Ritter et al. \shortcite{ritter2011data} who treat generation of conversational dialog as a statistical machine translation (SMT) problem. 
Ritter et al. \shortcite{ritter2011data} represents a break with previous and contemporaneous dialog work that relies extensively on hand-coded rules, typically either building statistical  models on top of heuristic rules or templates \cite{levin2000stochastic,young2010hidden,walker2003trainable,pieraccini2009we,wang2011improving} 
or learning generation rules from a minimal set of authored rules or labels \cite{oh2000stochastic,ratnaparkhi2002trainable,banchs2012iris,ameixa2014luke,nio2014developing,chen2013empirical}. More recently \cite{wen-EtAl2015} have used a Long Short-Term Memory (LSTM) \cite{hochreiter1997long} to learn from unaligned data in order to reduce the heuristic space of sentence planning and surface realization. 

The SMT model proposed by Ritter et al., on the other hand, is end-to-end, purely data-driven, and contains no explicit model of dialog structure; the model learns to converse from human-to-human conversational corpora. 
Progress in SMT stemming from the use of neural language models \cite{sutskever2014sequence,Gao2014,bahdanau2014neural,luong2014addressing} has inspired efforts to extend these neural techniques to SMT-based conversational response generation.
Sordoni et al. \shortcite{Sordoni2015} augments Ritter et al. \shortcite{ritter2011data} by rescoring outputs using a \sts model conditioned on conversation history. 
Other researchers have recently used \sts to directly generate responses in an end-to-end fashion without relying on SMT phrase tables \cite{serban2015hierarchical,shang2015neural,vinyals2015neural}.
Serban et al. \shortcite{serban2015hierarchical} propose a hierarchical neural 
model aimed at capturing dependencies over an extended conversation history. 
Recent work by Li et al. \shortcite{li2015diversity} measures mutual information between message and response in order to reduce the proportion of generic responses typical of \sts systems. Yao et al. \shortcite{YaoZP15} employ an intention network to maintain the relevance of responses.

Modeling of users and speakers has been extensively studied within the standard dialog modeling framework
(e.g., \cite{wahlster1989user,kobsa1990user,schatztnann2005effects,lin2011all}).
Since generating meaningful responses in an open-domain scenario is intrinsically difficult in conventional dialog systems, existing models often focus on generalizing character style on the basis of qualitative statistical analysis \cite{walker2012annotated,walker2011perceived}. 
The present work, by contrast, is in the vein of the \sts models of Vinyals and Le \shortcite{vinyals2015neural} and Li et al. \shortcite{li2015diversity}, enriching these models by training persona vectors directly from conversational data and relevant side-information, and incorporating these directly into the decoder.

\section{Sequence-to-Sequence Models}
\label{sec:seq2seq}

Given a sequence of inputs  $X=\{x_1,x_2,...,x_{n_X}\}$, an LSTM associates each time step with an input gate, a memory gate and an output gate, 
respectively denoted as $i_t$, $f_t$ and $o_t$.
We distinguish $e$ and $h$ where $e_t$ denotes the vector for an individual text unit (for example, a word or sentence) at time step $t$ while $h_t$ denotes the vector computed by the LSTM model at time $t$ by combining $e_t$ and $h_{t-1}$. 
$c_t$ is the cell state vector at time $t$, and
$\sigma$ denotes the sigmoid function. Then, the vector representation $h_t$ for each time step $t$ is given by:
\begin{equation}
\left[
\begin{array}{lr}
i_t\\
f_t\\
o_t\\
l_t\\
\end{array}
\right] =
\left[
\begin{array}{c}
\sigma\\
\sigma\\
\sigma\\
\text{tanh}\\
\end{array}
\right]
W\cdot
\left[
\begin{array}{c}
h_{t-1}\\
e_{t}^s\\
\end{array}
\right]
\end{equation}
\begin{equation}
c_t=f_t\cdot c_{t-1}+i_t\cdot l_t
\end{equation}
\begin{equation}
h_{t}^s=o_t\cdot \text{tanh}(c_t)
\end{equation}
where $W_i$, $W_f$, $W_o$, $W_l \in \mathbb{R}^{K\times 2K}$.
In \sts generation tasks, each input $X$ is paired with a sequence of outputs to predict: 
$Y=\{y_1,y_2,...,y_{n_Y}\}$. 
The LSTM defines a distribution over outputs and sequentially predicts tokens using a softmax function:
\begin{equation*}
\begin{aligned}
p(Y|X)
&=\prod_{t=1}^{n_y}p(y_t|x_1,x_2,...,x_t,y_1,y_2,...,y_{t-1})\\
&=\prod_{t=1}^{n_y}\frac{\exp(f(h_{t-1},e_{y_t}))}{\sum_{y'}\exp(f(h_{t-1},e_{y'}))}
\end{aligned}
\label{equ-lstm}
\end{equation*}
where $f(h_{t-1}, e_{y_t})$ denotes the activation function between $h_{t-1}$ and $e_{y_t}$.
Each sentence terminates with a special end-of-sentence symbol \eos. 
In keeping with common practices, inputs and outputs use different LSTMs with separate parameters to capture different compositional patterns.

During decoding, the algorithm terminates when an \eos token is predicted.
At each time step, either a greedy approach or beam search can be adopted for word prediction.

\section{Personalized Response Generation}
\label{sec:models}
\begin{figure*} [!ht]
\centering
\includegraphics[width=6in]{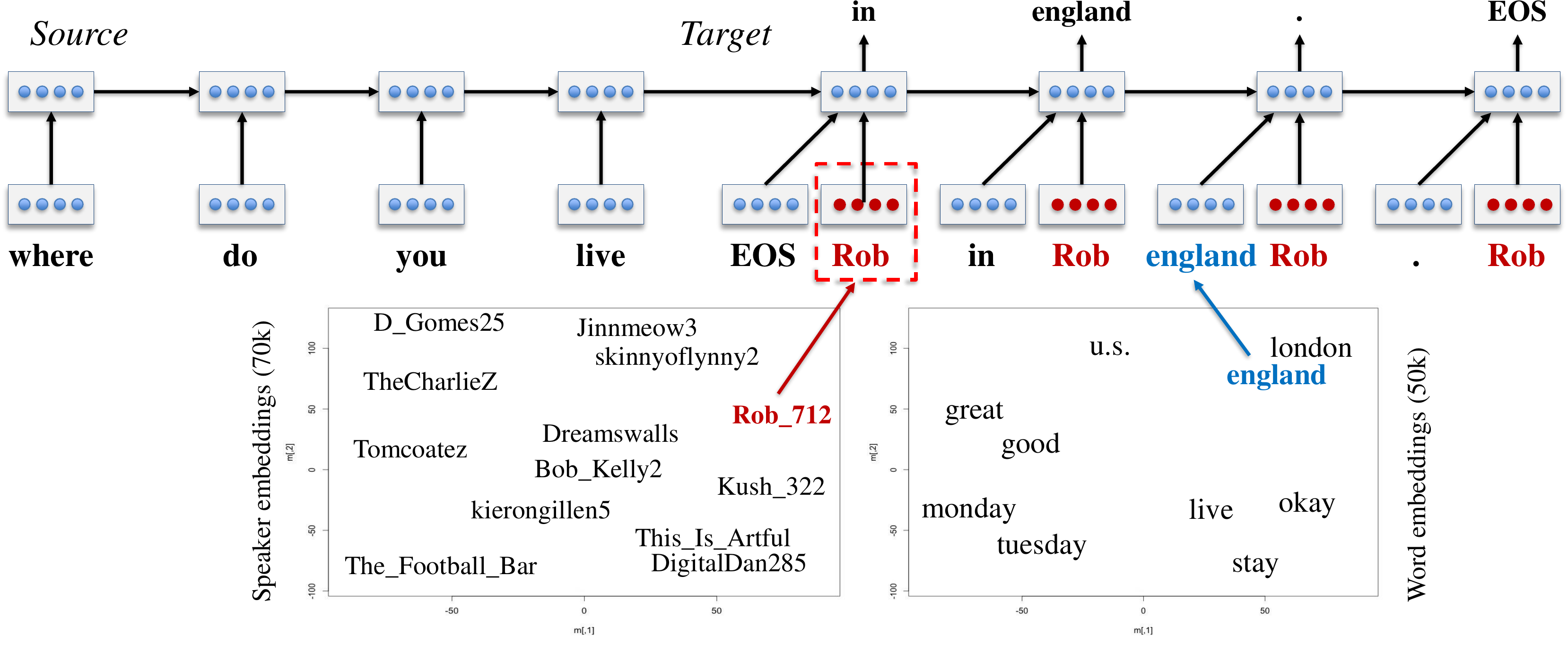}
\caption{Illustrative example of the Speaker Model introduced in this work. Speaker IDs close in embedding space tend to respond in the same manner. These speaker embeddings are learned jointly with word embeddings and all other parameters of the neural model via backpropagation. %
In this example, say Rob is a speaker clustered with people who often mention England in the training data, then the generation of the token `england' at time \mbox{$t=2$} would be much more likely than that of `u.s.'. A non-persona model would prefer generating {\it in the u.s.} if `u.s.' is more represented in the training data across all speakers.
}\label{fig1}
\end{figure*}

Our work introduces two persona-based models: the Speaker Model, 
which models the personality of the respondent, and the Speaker-Addressee Model which models the way the respondent adapts their speech to a given addressee --- a linguistic phenomenon known as lexical entrainment \cite{deutschpechmann82}. 

\subsection{Notation}
For the response generation task,
let $M$ denote the input word sequence (message) $M=\{m_1,m_2,...,m_I\}$. %
$R$ denotes the word sequence in response to $M$, where $R=\{r_1,r_2,...,r_J,$ \eos{}\} and $J$ is the length of the response (terminated by an \eos token). $r_t$ denotes a word token that is associated with a $K$ dimensional distinct word embedding $e_t$. $V$ is the vocabulary size.

\subsection{Speaker Model}
Our first model is the Speaker Model, which 
models the respondent alone.
This model represents each individual speaker as a vector or embedding, which encodes 
speaker-specific information (e.g., dialect, register, age, gender, personal information) that influences the content and style of her responses. Note that these attributes are not explicitly annotated, which would be tremendously expensive for our datasets. Instead, our model manages to cluster users along some of these traits (e.g., age, country of residence) based on the responses alone.

Figure \ref{fig1} gives a brief illustration of the Speaker Model. 
Each speaker $i\in [1,N]$ is associated with a user-level representation $v_i\in\mathbb{R}^{K\times 1}$. 
 As in standard \sts models, we first encode message  $S$ into a vector representation $h_S$ using the source LSTM. 
Then for each step in the target side, hidden units are obtained by 
combining the representation produced by the target LSTM at the previous time step, the word representations at the current time step, and the speaker embedding $v_i$:
\begin{equation}
\left[
\begin{array}{lr}
i_t\\
f_t\\
o_t\\
l_t\\
\end{array}
\right]=
\left[
\begin{array}{c}
\sigma\\
\sigma\\
\sigma\\
\text{tanh}\\
\end{array}
\right]
W\cdot
\left[
\begin{array}{c}
h_{t-1}\\
e_{t}^s\\
v_i\\
\end{array}
\right]
\end{equation}
\begin{equation}
c_t=f_t\cdot c_{t-1}+i_t\cdot l_t
\end{equation}
\begin{equation}
h_{t}^s=o_t\cdot \text{tanh}(c_t)
\end{equation}
where $W\in \mathbb{R}^{4K\times 3K}$. 
In this way, speaker information is encoded and 
injected into the hidden layer at each time step and thus helps predict personalized responses throughout the generation process.
The Speaker embedding $\{v_i\}$ is shared across all conversations that involve speaker $i$.  $\{v_i\}$ are learned by back propagating word prediction errors to each neural component during training. 

Another useful property of this model is that it 
helps {\it infer} answers to questions even if the evidence is not readily present in the training set.
This is important as
the training data does not contain explicit 
information about every
attribute of each user
(e.g., gender, age, country of residence).
The model learns speaker representations based on conversational content produced by different speakers, and speakers producing similar responses tend to have similar embeddings, occupying nearby positions in the vector space. 
This way, the training data of speakers nearby in vector space help increase the generalization capability of the
speaker model. For example, consider two speakers $i$ and $j$
who sound distinctly British, and who are therefore close in speaker 
embedding space. Now, suppose that, in the training data, speaker $i$ was asked {\it Where do you live?} and responded {\it in the UK}. Even if speaker $j$ was never asked the same question, this answer can help influence a good response from speaker $j$, and this without explicitly labeled geo-location information.
\begin{comment}
When a speaker is asked a questions like {\it Which company do you work for?} or {\it Where do you live?}, the evidence of which is not contained in the training set, the model can benefit from speakers that the current speaker resemble, take advantage of information of them and make inference. 
 
  the hometown of the speaker, the company he works for or his/her food preference for dinner for most of users as long as they don't explicitly mention these aspects.  But the model learns 
\end{comment}
  
\subsection{Speaker-Addressee Model}
A natural extension of the Speaker Model is a model that is sensitive to speaker-addressee interaction patterns within the conversation. Indeed,
speaking style, register, and content does not vary only with the identity of the speaker, but also with that of the addressee.
For example, in scripts for the TV series {\it Friends} used in some of our experiments, the character Ross often 
talks differently to his sister Monica than to Rachel,
with whom he is engaged in an on-again off-again relationship throughout the series. 
%
%
%
\begin{comment}
Modeling speaker-to-speaker interacting patterns requires that both of the speakers are involved in a long-term interactions with each other, from which their interacting patterns can be learned. 
\end{comment}

%
The proposed Speaker-Addressee Model operates as follows:
We wish to predict how speaker $i$ would respond to a message produced by speaker $j$. Similarly to the Speaker model, we associate each speaker with a $K$ dimensional speaker-level representation, namely $v_i$ for user $i$ and $v_j$ for user $j$. 
We obtain an interactive representation $V_{i,j}\in \mathbb{R}^{K\times 1}$ by linearly combining user vectors $v_i$ and $v_j$
in an attempt to model the interactive style of user $i$ towards user $j$,
\begin{equation}
V_{i,j}=\text{tanh}(W_1\cdot v_i+W_2\cdot v_2)
\end{equation}
where $W_1, W_2\in \mathbb{R}^{K\times K}$. 
$V_{i,j}$  is then linearly incorporated into LSTM models at each step in the target: 
\begin{equation}
\left[
\begin{array}{lr}
i_t\\
f_t\\
o_t\\
l_t\\
\end{array}
\right]=
\left[
\begin{array}{c}
\sigma\\
\sigma\\
\sigma\\
\text{tanh}\\
\end{array}
\right]
W\cdot
\left[
\begin{array}{c}
h_{t-1}\\
e_{t}^s\\
V_{i,j}\\
\end{array}
\right]
\end{equation}
\begin{equation}
c_t=f_t\cdot c_{t-1}+i_t\cdot l_t\\
\end{equation}
\begin{equation}
h_{t}^s=o_t\cdot \text{tanh}(c_t)
\end{equation}
$V_{i,j}$ 
depends on both speaker and addressee and
the same speaker will thus respond differently to a message from different interlocutors. 
One potential issue with Speaker-Addressee modelling is the difficulty involved in collecting a large-scale training dataset in which each speaker 
is involved in conversation with a wide variety of people. Like the Speaker Model, however, the Speaker-Addressee Model derives generalization capabilities from speaker embeddings.
Even if the two speakers
at test time ($i$ and $j$) were never involved in the same conversation in the training data, two speakers $i'$ and $j'$ who are respectively close in embeddings may have been, and this can help  modelling how $i$ should respond to $j$. 

\subsection{Decoding and Reranking}
For decoding, 
the N-best lists  are generated using the decoder with beam size \mbox{$B=200$}.
We set a maximum length of 20 for the generated candidates. 
Decoding operates as follows: At each time step, 
we first examine all \mbox{$B\times B$} possible next-word candidates, and add all hypothesis ending with an \eos token to the N-best list. We then preserve the top-$B$ unfinished hypotheses and move to the next word position. 

To deal with the issue that \sts models tend to generate generic and commonplace responses such as {\it I don't know}, we follow \newcite{li2015diversity} by reranking the generated N-best list 
using a scoring function that linearly combines 
 a length penalty and the log likelihood of the source given the target:
\begin{equation}
\log p(R|M,v)+\lambda\log p(M|R)+\gamma |R| 
\end{equation}
where $p(R|M,v)$ denotes the probability of the generated response given the message $M$ and the  respondent's speaker ID. 
$|R|$ denotes the length of the target and $\gamma$ denotes the associated penalty weight. We optimize $\gamma$ and $\lambda$ on N-best lists of response candidates generated from the development set using MERT \cite{mert} by optimizing \bleu.
To compute $p(M|R)$, 
we train an inverse \sts model by swapping messages and responses. We trained standard \sts models for $p(M|R)$  with no speaker information considered.

\section{Datasets}
\label{sec:data}
\subsection{Twitter Persona Dataset} 
\paragraph{Data Collection}
Training data for the {Speaker Model} was extracted from the Twitter FireHose for the six-month period beginning January 1, 2012.
We limited the sequences to those where the responders had engaged in at least 60 (and at most 300) 3-turn conversational interactions during the period, in other words, users who reasonably frequently engaged in conversation. This yielded a set of 74,003 users who took part in a minimum of 60 and a maximum of 164 conversational turns (average: 92.24, median:  90). 
The dataset extracted using responses by these ``conversationalists'' contained 24,725,711 3-turn sliding-window (context-message-response) conversational 
sequences.

In addition, we sampled 12000 3-turn conversations from the same user set from the Twitter FireHose for the three-month period beginning July 1, 2012, and set these aside as development, validation, and test sets (4000 conversational sequences each). Note that development, validation, and test sets for this data are single-reference, which is by design. Multiple reference responses would typically require acquiring responses from different people, which would confound different personas.

\paragraph{Training Protocols} We trained four-layer \sts models on the Twitter corpus following the approach of \cite{sutskever2014sequence}.
Details are as follows: 
\begin{itemize}
\reduceVerticalSpace
\item 4 layer LSTM models with 1,000 hidden cells for each layer.
\item Batch size is set to 128.
\item Learning rate is set to 1.0.
\item Parameters are initialized by sampling from the uniform distribution $[-0.1,0.1]$.
\item Gradients are clipped to avoid gradient explosion with a threshold of 5.
\item Vocabulary size is limited to 50,000.
\item Dropout rate is set to 0.2.
\end{itemize}
Source and target LSTMs use different sets of parameters.
We ran 14 epochs, and 
training took roughly a month to finish on a Tesla K40 GPU machine. 

As only speaker IDs of responses were specified when compiling the Twitter dataset, experiments on this dataset were limited to the {Speaker Model}.  

\subsection{Twitter Sordoni Dataset} 

The Twitter Persona Dataset was collected for this paper for experiments with speaker ID information. 
To obtain a point of 
comparison with prior state-of-the-art work \cite{Sordoni2015,li2015diversity}, 
we measure our baseline (non-persona) LSTM model against prior 
work on the dataset of \cite{Sordoni2015}, which we call the Twitter Sordoni Dataset. 
We only use
its test-set portion, which contains
responses for 2114 context and messages. 
It is important to note that 
the Sordoni dataset offers up to 10 references per message, while the 
Twitter Persona dataset has only 1 reference per message. Thus \bleu scores cannot be compared across the two Twitter datasets (\bleu scores on 10 references 
are generally much higher than with 1 reference).
Details of this dataset are in \cite{Sordoni2015}.

\subsection{Television Series Transcripts} 
\paragraph{Data Collection} For the dyadic Speaker-Addressee Model we used scripts from the American television comedies {\it Friends}\footnote{\url{https://en.wikipedia.org/wiki/Friends}} and {\it The Big Bang Theory},\footnote{\url{https://en.wikipedia.org/wiki/The_Big_Bang_Theory}} available from Internet Movie Script Database (IMSDb).\footnote{\url{http://www.imsdb.com}}
We collected 13 main characters from the two series in a corpus of 69,565 turns. 
We split the corpus into training/development/testing sets, with development and testing sets each of about 2,000 turns. 

\paragraph{Training} 
Since the relatively small size of the dataset does not allow for training an open domain dialog model, we adopted a domain adaption strategy where we first trained a standard \sts models using a much larger OpenSubtitles (OSDb) dataset \cite{tiedemann2009news}, and then adapting the pre-trained model to the TV series dataset. 

The OSDb dataset is a large, noisy, open-domain dataset containing roughly 60M-70M scripted lines spoken by movie characters. 
This dataset does not specify which character speaks each subtitle line, which prevents us from inferring speaker turns. 
Following Vinyals et al. (2015), we make the simplifying assumption that each line of subtitle constitutes a full speaker turn.\footnote{This introduces a degree of noise as consecutive lines are not necessarily from the same scene or two different speakers.}
We trained standard \sts models on OSDb dataset, following the protocols already described in Section 5.1. 
We run 10 iterations over the training set.

We initialize word embeddings and LSTM parameters in the Speaker Model and the Speaker-Addressee model using parameters learned from OpenSubtitles datasets. 
User embeddings are randomly initialized  from $[-0.1,0.1]$. 
We then ran 5 additional epochs until the perplexity on the development set stabilized.

\section{Experiments}
\label{sec:experiments}
\subsection{Evaluation}
Following \cite{Sordoni2015,li2015diversity}
we used \bleu \cite{Papineni2002BLEU} 
for parameter tuning and evaluation. 
\bleu has been shown to correlate well with human judgment on the response generation task, as demonstrated in \cite{galley2015deltableu}. 
%
\begin{comment}
\cite{li2015diversity} also adopted  {\it distinct-1} and {\it distinct-2} to calculating the number of distinct unigrams and bigrams in generated responses 
to measure the degree of diversity. The value is scaled by total number of generated tokens to avoid favoring long sentences. 
We also report the degree of diversity following the protocols defined in \cite{li2015diversity}.  
\end{comment}
Besides \bleu scores, we also report perplexity as an indicator of model capability.

\begin{table}
\centering
\begin{tabular}{lc}
System                                    & \bleu \\ \hline
MT baseline \cite{ritter2011data}         & 3.60\% \\ \hline
Standard LSTM MMI \cite{li2015diversity}  & 5.26\% \\
Standard LSTM MMI (our system)            & 5.82\% \\ \hline
{\it Human}                               & {\it 6.08\%}\\
\end{tabular}
\caption{\bleu on the Twitter Sordoni dataset (10 references). We contrast our baseline against an SMT baseline \cite{ritter2011data}, and the best result \cite{li2015diversity} on the established
dataset of \cite{Sordoni2015}.
The last result is for a human oracle, but it is not directly comparable as the oracle \bleu is computed in a leave-one-out fashion, having one less reference available. We nevertheless provide
this result to give a sense that these \bleu scores of 5-6\% are not unreasonable.}
\label{twitter-baselines}
\end{table}

\subsection{Baseline}
Since our main experiments are with a new dataset (the Twitter Persona Dataset), we first show that our LSTM baseline is competitive with the state-of-the-art \cite{li2015diversity} on an established
dataset, the Twitter Sordoni Dataset \cite{Sordoni2015}.
Our baseline is simply our implementation of the LSTM-MMI of \cite{li2015diversity}, so results should be relatively close to their reported results.
Table~\ref{twitter-baselines} summarizes our results against prior work.
\begin{comment}
The comparison is particularly important because the
test set of \cite{Sordoni2015,li2015diversity} offers up to 10 references, while ours is single-reference (a multi-reference test set, would involve different respondents and confound different personas, which we want to avoid).
\end{comment}
We see that our system actually does better than \cite{li2015diversity}, and
we attribute the improvement to a larger training corpus, the use of dropout during training, and possibly to the ``conversationalist'' nature of our corpus.

\begin{table}
\centering
\begin{tabular}{ccc}
Model&Standard LSTM&Speaker Model \\\hline
Perplexity&47.2&42.2 ($-10.6\%$) \\
\end{tabular}
\caption{Perplexity for standard \sts and the Speaker model 
on the Twitter Persona development set.}
\label{twitter-per}
\end{table}
\begin{table}
\centering
\begin{tabular}{lll}
Model&Objective& \bleu \\\hline
Standard LSTM &MLE& 0.92\% \\
Speaker Model & MLE&1.12\% (+21.7$\%$) \\\hline
Standard LSTM &MMI& 1.41\% \\
Speaker Model & MMI&1.66\% (+11.7$\%$) \\
\end{tabular}
\caption{
\bleu on the Twitter Persona dataset (1 reference), for the
standard \sts model and the Speaker model using as objective either maximum likelihood (MLE) or maximum mutual information (MMI).}
\label{twitter-bleu}
\end{table}

\begin{table*}
\centering
\begin{tabular}{cccc}
Model&Standard LSTM&Speaker Model& Speaker-Addressee Model \\\hline
Perplexity&27.3&25.4 ($-7.0\%$)& 25.0 ($-8.4\%$)  \\
\end{tabular}
\caption{Perplexity for standard \sts and persona models on the TV series dataset.}
\label{tv-per}
\end{table*}
\begin{table*}
\centering
\begin{tabular}{ccccc}
Model&Standard LSTM&Speaker Model& Speaker-Addressee Model \\\hline
MLE&1.60\%& 1.82\% ($+13.7\%$)& 1.83\%  ($+14.3\%$) \\
MMI&1.70\%& 1.90\% ($+10.6\%$) &1.88\% ($+10.9\%$) \\\hline
\end{tabular}
\caption{
\bleu on the TV series dataset (1 reference), for the
standard \sts and persona models.}
\label{tv-bleu}
\end{table*}

\subsection{Results}
We first report performance on the Twitter Persona dataset.
Perplexity is reported in Table \ref{twitter-per}. We observe about a $10\%$ decrease in perplexity for the Speaker model over the standard \sts model. 
In terms of \bleu scores (Table~\ref{twitter-bleu}), a significant performance boost 
is observed for 
 the Speaker model over the standard \sts model, yielding an increase of $21\%$
in the maximum likelihood (MLE) setting and $11.7\%$ for mutual information setting (MMI). 
In line with findings in \cite{li2015diversity}, we observe a consistent performance boost introduced by 
the MMI objective function 
over a standard \sts model based on the MLE objective function. 
It is worth noting that our persona models are more beneficial to the MLE models
than to the MMI models. This result is intuitive as the persona models help make 
Standard LSTM MLE outputs more informative and less bland, and thus make the use 
of MMI less critical.
\begin{comment}
It is noting that the difference between MLE persona models and MLE \sts models is greater than that between MMI persona model and MMI \sts models. 
Such a result is intuitive since both persona models and \sts models use the same $p(M|R)$ for later MMI reranking, which is obtained without user-level information. Such a reranking process would dilute the discrepancy between the persona models and vanilla \sts models. 
\end{comment}

For the TV Series dataset,  perplexity and \bleu  scores are respectively  reported in Table \ref{tv-per} and Table \ref{tv-bleu}.
As can be seen, the Speaker and Speaker-Addressee models respectively achieve perplexity values of 25.4 and 25.0 on the TV-series dataset, $7.0\%$ and $8.4\%$ percent lower than the correspondent standard \sts models. 
In terms of \bleu score,  we observe a similar performance  boost as on the Twitter dataset, in which the Speaker model
and the Speaker-Addressee model 
 outperform the standard \sts model by $13.7\%$ and $10.6\%$.
By comparing the Speaker-Addressee model against the Speaker model on the TV Series dataset, we do not observe a significant difference.  
We suspect that this is primarily due to the relatively small size of the dataset where the interactive patterns might not be fully captured. 
Smaller values of perplexity are observed for the Television Series dataset than the Twitter dataset, the perplexity of which is over 40, presumably due to the noisier nature of Twitter dialogues. 
\subsection{Qualitative Analysis}

\begin{table}
\setlength{\tabcolsep}{4pt}
\center
{\small
\begin{tabularx}{\linewidth}{rl} \hline
\Message   & Where is your hometown?\\\hline
{\it baseline}    & I was born in Canada.\\\hline
\User{1}   & I'm from England.\\
\User{2}   & I'm from Manchester.\\
\User{3}   & I'm from Liverpool.\\
\User{4}   & England. You?\\
\User{5}   & My hometown.\\
\User{6}   & I'm from Texas.\\
\User{7}   & I'm from LA.\\
\User{8}   & England.\\
\User{9}   & I was born here.\\
\User{10}  & I was born in the us.\\\hline\hline
\Message   & What company do you work for?\\\hline
{\it baseline}    & I work for a company.\\\hline
\User{1}   & I don't have a job.\\
\User{2}   & I'm a manager.\\
\User{3}   & I'm working. {\color{red} $\#$}\\
\User{4}   & I work for a company. \\
\User{5}   & I'm working. {\color{red} $\#$}\\
\User{6}   & Customer service.\\
\User{7}   & I work in retail.\\
\User{8}   & I work in a cafe.\\
\User{9}   & I work at a gas station.\\
\User{10}  & I don't know {\color{red} $\#$} \\\hline\hline
\Message   & What did you have for dinner?\\\hline
{\it baseline}    & I had fish and chips.\\\hline
\User{1}   & I had spag bol.\\
\User{2}   & Chicken and chips.\\
\User{3}   & Chicken and rice.\\
\User{4}   & Fish and chips.\\
\User{5}   & I had spag bol.\\
\User{6}   & I had Mexican food.\\
\User{7}   & Salad...\\
\User{8}   & I had chicken and chips.\\
\User{9}   & I had spag bol.\\
\User{10}  & Pizza.\\\hline
\end{tabularx}
}
\caption{Responses generated by the baseline (LSTM-MMI) and the Speaker Model for ten randomly selected users, without cherry picking. {\color{red} $\#$} indicates poor-quality responses produced by the system.}
\label{example1}
\end{table}

\paragraph{Diverse Responses by Different Speakers}
Table \ref{example1} represents responses generated by persona models in response to three different input questions. We randomly selected 10 speakers (without cherry-picking) from the original Twitter dataset. We collected their user level representations from a speaker look-up table and integrated them into the decoding models.  The model tends to generate specific responses for different people in response to the factual questions.\footnote{There appears to be a population bias in the training set that favors British users.}  

Table \ref{addressees} shows responses generated from the {\it Speaker-Addressee Model} using the TV-series dataset. Interestingly, we regularly observe 
that this model is sensitive to the identity of the addressee, generating words specifically targeted at that addressee (e.g., her name).
For example, the model produces {\it Of course, I love you, Emily} in response to an input from {\it Emily}. Also, the model generates {\it Of course I love you. ( kisses him)}, where the pronoun ``him'' accurately identifies the gender of the addressee.

\paragraph{Human Evaluation} We conducted a human evaluation of outputs from the Speaker Model, using 
a crowdsourcing service. 
Since we cannot expect crowdsourced human judges to know or attempt to learn the ground truth of Twitter users who are not well-known public figures, we designed our experiment to evaluate the consistency of outputs associated with the speaker IDs. To this end, we collected 24 pairs of questions for which we would expect responses to be consistent if the persona model is coherent.  For example, responses to the questions {\it What country do you live in?} and {\it What city do you live in?} would be considered consistent if the answers were {\it England} and {\it London} respectively, but not if they were {\it UK} and {\it Chicago}.  Similarly, the responses to {\it Are you vegan or vegetarian?} and {\it Do you eat beef?} are consistent if the answers generated are {\it vegan} and {\it absolutely not}, but not if they are {\it vegan} and {\it I love beef}.  We collected 20 pairs of outputs for randomly-selected personas provided by the Speaker Model for each question pair (480 response pairs total). We also obtained the corresponding outputs from the baseline MMI-enhanced \sts system. 

\begin{table}
\setlength{\tabcolsep}{4pt}
\center
{\small
\begin{tabularx}{\linewidth}{rl} \hline
\Message & Do you love me?\\\hline
{\it baseline} & Of course, I love you.  \\\hline
& {\it addressee: {\it Emily}, speaker: Sheldon}\\
{\it response} & Of course, I love you, Emily.\\\hline
& {\it addressee: {\it Emily}, speaker: Penny}\\
{\it response} & Of course, I love you, Emily.\\\hline
& {\it addressee: {\it Leonard}, speaker: Penny}\\
{\it response} & Of course, I love you, Leonard.\\\hline
& {\it addressee: {\it Monica}, speaker: Chandler}\\
{\it response} & Of course I love you. I want to marry you.\\\hline
& {\it addressee: {\it  Rachel}, speaker: Chandler}\\
{\it response} & Of course I love you.\\\hline
& {\it addressee: {\it  Ross}, speaker: Chandler}\\
{\it response} & Of course I love you.\\\hline
& {\it addressee: {\it  Ross}, speaker: Rachel}\\
{\it response} & Of course I love you. (kisses him) \\\hline
& {\it addressee: {\it  Emily}, speaker: Rachel}\\
{\it response} & Of course I love you. \\\hline
\end{tabularx}
}
\caption{Responses to {\it Do you love me?} from the {\it Speaker-Addressee} model on the TV-series dataset using different addressees and speakers.}
\label{addressees}
\end{table}

\begin{table}
\setlength{\tabcolsep}{4pt}
\center
{\small
\begin{tabularx}{\linewidth}{rl} \hline
{\bf User1} &\\\hline
\Message    & Where is your hometown?\\
\Response   & I'm from England.\\
\Message    & Where are you from?\\
\Response   & I'm from England.\\
\Message    & In which city do you live now?\\
\Response   & I live in London.\\
\Message    & In which country do you live now?\\
\Response   & I live in England.\\\hline
\Message    & What is your major?\\
\Response   & Business. You?\\
\Message    & What did you study in college?\\
\Response   & I did business studies.\\\hline
\Message    & How old are you?\\
\Response   & I'm 18.\\
\Message    & What is your age?\\
\Response   & I'm 18.\\\hline\hline
{\bf User2} \\\hline
\Message    & Where is your hometown? \\
\Response   & I live in Indonesia.\\
\Message    & Where are you from?\\
\Response   & England, you?\\
\Message    & In which city do you live now?\\
\Response   & I live in Jakarta.\\
\Message    & In which country do you live now?\\
\Response   & I live in Indonesia.\\\hline
\Message    & What is your major?\\
\Response   & Business, you?\\
\Message    & What did you study in college?\\
\Response   & Psychology, you?\\\hline
\Message    & How old are you?\\
\Response   & I'm 18.\\
\Message    & What is your age?\\
\Response   & I'm 16.\\\hline%
\end{tabularx}
}
\caption{Examples of speaker consistency and inconsistency generated by the Speaker Model}
\label{example2}
\end{table}

Since our purpose is to measure the gain in consistency over the baseline system, we presented the pairs of answers system-pairwise, i.e., 4 responses, 2 from each system,  displayed on the screen, and asked judges to decide which of the two systems was more consistent.  The position in which the system pairs were presented on the screen was randomized.  
The two systems were judged on 5-point zero-sum scale, assigning a score of 2 (-2) if one system was judged more (less) consistent than the other, and 1 (-1) if one was rated ``somewhat'' more (less) consistent.  Ties were assigned a score of zero. Five judges rated each pair and their scores were averaged and remapped into 5 equal-width bins. After discarding ties, we found the persona model was judged either ``more consistent'' or ``somewhat more consistent'' in 56.7\% of cases. If we ignore the ``somewhat more consistent'' judgments, the persona model wins in 6.1\% of cases, compared with only 1.6\% for the baseline model. 
It should be emphasized that the baseline model is a strong baseline, 
%
\begin{comment}
The two systems were judged on a 5-point scale, assigning a score of 2 (respectively -2) if the persona system was judged much more (respectively less) consistent than the baseline, 1 (respectively -1) if ``mostly'' more (respectively less) consistent, and 0 otherwise. 
Five judges rated each pair, and scores were averaged across judges.\footnote{To turn this average back into a 5-point scale, we mapped average scores into 5 equal-size bins.}
%
%
After removing ties
%
we found that
the persona model (respectively baseline) was judged either ``more consistent'' or ``somewhat more consistent'' than the baseline (respectively persona model) in 56.7\% (respectively 43.3\%) of the cases. Ignoring ``somewhat more consistent" judgments, the persona model is more consistent in 6.0\% of the cases, while the baseline only 1.6\% of the time.
It should be stressed that the latter is a strong baseline, 
\end{comment}
since it represents the consensus of all 70K Twitter users in the dataset\footnote{{\it I'm not pregnant} is an excellent consensus answer to the question {\it Are you pregnant?}, while {\it I'm pregnant} is consistent as a response only in the case of someone who also answers the question {\it Are you a guy or a girl?} with something in the vein of {\it I'm a girl}.}.

Table \ref{example2} illustrates how consistency is an emergent property of two arbitrarily selected users. The model is capable of discovering the relations between different categories of location such as London and the UK, Jakarta and Indonesia. However, the model also makes inconsistent response decisions, generating different answers in the second example in response to questions asking about age or major. 
Our proposed persona models integrate user embeddings into the LSTM, and thus can be viewed as encapsulating a trade-off between a persona-specific generation model and a general conversational model. 

\section{Conclusions}
\label{sec:conclusion}
We have presented two persona-based response generation models for open-domain conversation generation. 
There are many other dimensions of speaker behavior, such as mood and emotion, that 
are beyond the scope of the current paper and 
must be left to future work. 

Although the gains presented by our new models are not spectacular, the systems outperform our baseline \sts systems in terms of \bleu, perplexity, and human judgments of speaker consistency. 
We have demonstrated that by encoding personas in distributed representations, we are able
to capture personal characteristics such as speaking style and background information. 
In the Speaker-Addressee model, moreover, the evidence suggests that there is benefit in capturing dyadic interactions. 

Our ultimate goal is to be able to take the profile of an arbitrary individual whose identity is not known in advance, and generate conversations that accurately emulate that individual's persona in terms of linguistic response behavior and other salient characteristics. 
Such a capability will dramatically change the ways in which we interact with dialog agents of all kinds, opening up rich new possibilities for user interfaces. 
Given a sufficiently large training corpus in which a sufficiently rich variety of speakers is represented, this objective does not seem too far-fetched.

\section*{Acknowledgments}

We with to thank Stephanie Lukin, Pushmeet Kohli, Chris Quirk, Alan Ritter, and Dan Jurafsky for helpful discussions.

\bibliographystyle{acl2016}
\bibliography{person_response-clean}  

\begin{thebibliography}{}

\bibitem[\protect\citename{Ameixa \bgroup et al.\egroup }2014]{ameixa2014luke}
David Ameixa, Luisa Coheur, Pedro Fialho, and Paulo Quaresma.
\newblock 2014.
\newblock Luke, {I} am your father: dealing with out-of-domain requests by
  using movies subtitles.
\newblock In {\em Intelligent Virtual Agents}, pages 13--21. Springer.

\bibitem[\protect\citename{Bahdanau \bgroup et al.\egroup
  }2015]{bahdanau2014neural}
Dzmitry Bahdanau, Kyunghyun Cho, and Yoshua Bengio.
\newblock 2015.
\newblock Neural machine translation by jointly learning to align and
  translate.
\newblock In {\em Proc. of the International Conference on Learning
  Representations (ICLR)}.

\bibitem[\protect\citename{Banchs and Li}2012]{banchs2012iris}
Rafael~E Banchs and Haizhou Li.
\newblock 2012.
\newblock {IRIS}: a chat-oriented dialogue system based on the vector space
  model.
\newblock In {\em Proc. of the ACL 2012 System Demonstrations}, pages 37--42.

\bibitem[\protect\citename{Chen \bgroup et al.\egroup }2013]{chen2013empirical}
Yun-Nung Chen, Wei~Yu Wang, and Alexander Rudnicky.
\newblock 2013.
\newblock An empirical investigation of sparse log-linear models for improved
  dialogue act classification.
\newblock In {\em Acoustics, Speech and Signal Processing (ICASSP), 2013 IEEE
  International Conference on}, pages 8317--8321. IEEE.

\bibitem[\protect\citename{Deutsch and Pechmann}1982]{deutschpechmann82}
Werner Deutsch and Thomas Pechmann.
\newblock 1982.
\newblock Social interaction and the development of definite descriptions.
\newblock {\em Cognition}, 11:159--184.

\bibitem[\protect\citename{Galley \bgroup et al.\egroup
  }2015]{galley2015deltableu}
Michel Galley, Chris Brockett, Alessandro Sordoni, Yangfeng Ji, Michael Auli,
  Chris Quirk, Margaret Mitchell, Jianfeng Gao, and Bill Dolan.
\newblock 2015.
\newblock \dbleu: A discriminative metric for generation tasks with
  intrinsically diverse targets.
\newblock In {\em Proc. of ACL-IJCNLP}, pages 445--450, Beijing, China, July.

\bibitem[\protect\citename{Gao \bgroup et al.\egroup }2014]{Gao2014}
Jianfeng Gao, Xiaodong He, Wen-tau Yih, and Li~Deng.
\newblock 2014.
\newblock Learning continuous phrase representations for translation modeling.
\newblock In {\em Proc. of ACL}, pages 699--709, Baltimore, Maryland.

\bibitem[\protect\citename{Hochreiter and Schmidhuber}1997]{hochreiter1997long}
Sepp Hochreiter and J{\"u}rgen Schmidhuber.
\newblock 1997.
\newblock Long short-term memory.
\newblock {\em Neural computation}, 9(8):1735--1780.

\bibitem[\protect\citename{Kobsa}1990]{kobsa1990user}
Alfred Kobsa.
\newblock 1990.
\newblock User modeling in dialog systems: Potentials and hazards.
\newblock {\em AI \& society}, 4(3):214--231.

\bibitem[\protect\citename{Levin \bgroup et al.\egroup
  }2000]{levin2000stochastic}
Esther Levin, Roberto Pieraccini, and Wieland Eckert.
\newblock 2000.
\newblock A stochastic model of human-machine interaction for learning dialog
  strategies.
\newblock {\em IEEE Transactions on Speech and Audio Processing}, 8(1):11--23.

\bibitem[\protect\citename{Li \bgroup et al.\egroup }2016]{li2015diversity}
Jiwei Li, Michel Galley, Chris Brockett, Jianfeng Gao, and Bill Dolan.
\newblock 2016.
\newblock A diversity-promoting objective function for neural conversation
  models.
\newblock In {\em Proc. of NAACL-HLT}.

\bibitem[\protect\citename{Lin and Walker}2011]{lin2011all}
Grace~I Lin and Marilyn~A Walker.
\newblock 2011.
\newblock All the world's a stage: Learning character models from film.
\newblock In {\em Proceedings of the Seventh {AAAI} Conference on Artificial
  Intelligence and Interactive Digital Entertainment {(AIIDE)}}.

\bibitem[\protect\citename{Luong \bgroup et al.\egroup
  }2015]{luong2014addressing}
Thang Luong, Ilya Sutskever, Quoc Le, Oriol Vinyals, and Wojciech Zaremba.
\newblock 2015.
\newblock Addressing the rare word problem in neural machine translation.
\newblock In {\em Proc. of ACL}, pages 11--19, Beijing, China, July.

\bibitem[\protect\citename{Nio \bgroup et al.\egroup }2014]{nio2014developing}
Lasguido Nio, Sakriani Sakti, Graham Neubig, Tomoki Toda, Mirna Adriani, and
  Satoshi Nakamura.
\newblock 2014.
\newblock Developing non-goal dialog system based on examples of drama
  television.
\newblock In {\em Natural Interaction with Robots, Knowbots and Smartphones},
  pages 355--361. Springer.

\bibitem[\protect\citename{Och}2003]{mert}
Franz~Josef Och.
\newblock 2003.
\newblock Minimum error rate training in statistical machine translation.
\newblock In {\em Proceedings of the 41st Annual Meeting of the Association for
  Computational Linguistics}, pages 160--167, Sapporo, Japan, July. Association
  for Computational Linguistics.

\bibitem[\protect\citename{Oh and Rudnicky}2000]{oh2000stochastic}
Alice~H Oh and Alexander~I Rudnicky.
\newblock 2000.
\newblock Stochastic language generation for spoken dialogue systems.
\newblock In {\em Proceedings of the 2000 ANLP/NAACL Workshop on Conversational
  systems-Volume 3}, pages 27--32.

\bibitem[\protect\citename{Papineni \bgroup et al.\egroup
  }2002]{Papineni2002BLEU}
Kishore Papineni, Salim Roukos, Todd Ward, and Wei-Jing Zhu.
\newblock 2002.
\newblock {\sc Bleu}: a method for automatic evaluation of machine translation.
\newblock In {\em Proc. of ACL}, pages 311--318.

\bibitem[\protect\citename{Pieraccini \bgroup et al.\egroup
  }2009]{pieraccini2009we}
Roberto Pieraccini, David Suendermann, Krishna Dayanidhi, and Jackson Liscombe.
\newblock 2009.
\newblock Are we there yet? research in commercial spoken dialog systems.
\newblock In {\em Text, Speech and Dialogue}, pages 3--13. Springer.

\bibitem[\protect\citename{Ratnaparkhi}2002]{ratnaparkhi2002trainable}
Adwait Ratnaparkhi.
\newblock 2002.
\newblock Trainable approaches to surface natural language generation and their
  application to conversational dialog systems.
\newblock {\em Computer Speech \& Language}, 16(3):435--455.

\bibitem[\protect\citename{Ritter \bgroup et al.\egroup }2011]{ritter2011data}
Alan Ritter, Colin Cherry, and William~B Dolan.
\newblock 2011.
\newblock Data-driven response generation in social media.
\newblock In {\em Proceedings of the Conference on Empirical Methods in Natural
  Language Processing}, pages 583--593.

\bibitem[\protect\citename{Schatztnann \bgroup et al.\egroup
  }2005]{schatztnann2005effects}
Jost Schatztnann, Matthew~N Stuttle, Karl Weilhammer, and Steve Young.
\newblock 2005.
\newblock Effects of the user model on simulation-based learning of dialogue
  strategies.
\newblock In {\em Automatic Speech Recognition and Understanding, 2005 IEEE
  Workshop on}, pages 220--225.

\bibitem[\protect\citename{Serban \bgroup et al.\egroup
  }2015]{serban2015hierarchical}
Iulian~V Serban, Alessandro Sordoni, Yoshua Bengio, Aaron Courville, and Joelle
  Pineau.
\newblock 2015.
\newblock Building end-to-end dialogue systems using generative hierarchical
  neural network models.
\newblock In {\em Proc. of AAAI}.

\bibitem[\protect\citename{Shang \bgroup et al.\egroup }2015]{shang2015neural}
Lifeng Shang, Zhengdong Lu, and Hang Li.
\newblock 2015.
\newblock Neural responding machine for short-text conversation.
\newblock In {\em ACL-IJCNLP}, pages 1577--1586.

\bibitem[\protect\citename{Sordoni \bgroup et al.\egroup }2015]{Sordoni2015}
Alessandro Sordoni, Michel Galley, Michael Auli, Chris Brockett, Yangfeng Ji,
  Meg Mitchell, Jian-Yun Nie, Jianfeng Gao, and Bill Dolan.
\newblock 2015.
\newblock A neural network approach to context-sensitive generation of
  conversational responses.
\newblock In {\em Proc. of NAACL-HLT}.

\bibitem[\protect\citename{Sutskever \bgroup et al.\egroup
  }2014]{sutskever2014sequence}
Ilya Sutskever, Oriol Vinyals, and Quoc~V Le.
\newblock 2014.
\newblock Sequence to sequence learning with neural networks.
\newblock In {\em Advances in neural information processing systems (NIPS)},
  pages 3104--3112.

\bibitem[\protect\citename{Tiedemann}2009]{tiedemann2009news}
J{\"o}rg Tiedemann.
\newblock 2009.
\newblock News from {OPUS} -- a collection of multilingual parallel corpora
  with tools and interfaces.
\newblock In {\em Recent advances in natural language processing}, volume~5,
  pages 237--248.

\bibitem[\protect\citename{Vinyals and Le}2015]{vinyals2015neural}
Oriol Vinyals and Quoc Le.
\newblock 2015.
\newblock A neural conversational model.
\newblock In {\em Proc. of {ICML} Deep Learning Workshop}.

\bibitem[\protect\citename{Wahlster and Kobsa}1989]{wahlster1989user}
Wolfgang Wahlster and Alfred Kobsa.
\newblock 1989.
\newblock {\em User models in dialog systems}.
\newblock Springer.

\bibitem[\protect\citename{Walker \bgroup et al.\egroup
  }2003]{walker2003trainable}
Marilyn~A Walker, Rashmi Prasad, and Amanda Stent.
\newblock 2003.
\newblock A trainable generator for recommendations in multimodal dialog.
\newblock In {\em INTERSPEECH}.

\bibitem[\protect\citename{Walker \bgroup et al.\egroup
  }2011]{walker2011perceived}
Marilyn~A Walker, Ricky Grant, Jennifer Sawyer, Grace~I Lin, Noah
  Wardrip-Fruin, and Michael Buell.
\newblock 2011.
\newblock Perceived or not perceived: Film character models for expressive nlg.
\newblock In {\em Interactive Storytelling}, pages 109--121. Springer.

\bibitem[\protect\citename{Walker \bgroup et al.\egroup
  }2012]{walker2012annotated}
Marilyn~A Walker, Grace~I Lin, and Jennifer Sawyer.
\newblock 2012.
\newblock An annotated corpus of film dialogue for learning and characterizing
  character style.
\newblock In {\em LREC}, pages 1373--1378.

\bibitem[\protect\citename{Wang \bgroup et al.\egroup }2011]{wang2011improving}
William~Yang Wang, Ron Artstein, Anton Leuski, and David Traum.
\newblock 2011.
\newblock Improving spoken dialogue understanding using phonetic mixture
  models.
\newblock In {\em FLAIRS Conference}.

\bibitem[\protect\citename{Wen \bgroup et al.\egroup }2015]{wen-EtAl2015}
Tsung-Hsien Wen, Milica Gasic, Nikola Mrk\v{s}i\'{c}, Pei-Hao Su, David
  Vandyke, and Steve Young.
\newblock 2015.
\newblock Semantically conditioned {LSTM}-based natural language generation for
  spoken dialogue systems.
\newblock In {\em Proc. of EMNLP}, pages 1711--1721, Lisbon, Portugal,
  September. Association for Computational Linguistics.

\bibitem[\protect\citename{Yao \bgroup et al.\egroup }2015]{YaoZP15}
Kaisheng Yao, Geoffrey Zweig, and Baolin Peng.
\newblock 2015.
\newblock Attention with intention for a neural network conversation model.
\newblock {\em CoRR}, abs/1510.08565.

\bibitem[\protect\citename{Young \bgroup et al.\egroup }2010]{young2010hidden}
Steve Young, Milica Ga{\v{s}}i{\'c}, Simon Keizer, Fran{\c{c}}ois Mairesse,
  Jost Schatzmann, Blaise Thomson, and Kai Yu.
\newblock 2010.
\newblock The hidden information state model: A practical framework for
  pomdp-based spoken dialogue management.
\newblock {\em Computer Speech \& Language}, 24(2):150--174.

\end{thebibliography}

\end{document}